\documentclass[10pt, twocolumn]{article}
\usepackage{geometry}
\usepackage{rahulbhadani}
\title{\color{red}{Path Planning of Unmanned System using Carrot-chasing Algorithm}}
\author{Rahul Bhadani\footnote{Email: rahulbhadani@email.arizona.edu}\\Department of Electrical \& Computer Engineering\\The University of Arizona}
\date{}

\usepackage{graphicx}
\usepackage{tgpagella}

\def\BibTeX{{\rm B\kern-.05em{\sc i\kern-.025em b}\kern-.08em
    T\kern-.1667em\lower.7ex\hbox{E}\kern-.125emX}}
\begin{document}

\maketitle
\begin{abstract}
When an unmanned system is launched for a mission-critical task, it is required to follow a predetermined path. It means the unmanned system requires a path following algorithm for the completion of the mission. Since the predetermined path is typically given by a set of data-points, not only the curvature and derivative of the pre-determined path are absent, but also it requires a large size of on-board memory. In this work, we study a simple path following algorithm called Carrot-chasing algorithm that uses a simple controller in the form of a proportional controller to control the movement of an unmanned system.
\end{abstract}

\section{Introduction}

Unmanned systems are mainly used by government and defense organizations but with the significant advancement in electronics and due to the availability of low-cost sensors, there has been wide interest in using low-cost unmanned systems among hobbyists, students, and instructors in universities and schools. These unmanned systems fall into several categories such as Unmanned Aerial Vehicles (UAV), Autonomous Ground Vehicles (AGV), and Unmanned Underwater Vehicles (UUV). Dynamics and Kinematics of these Unmanned Systems vary and their corresponding constraints vary depending upon the environment in which they operate. Nevertheless, the underlying principle of the Path following algorithm is the same in all cases and mostly differ in terms of precision and complexity. For the work presented in this paper, it has been assumed that vehicle is a point mass and their movement is restricted in the x-y plane. Further, it is also assumed that there is no disturbance and noise from the outer environment.

\section{Problem Formulation}

\begin{figure}[htpb]
 \centering
\includegraphics[width=0.8\linewidth]{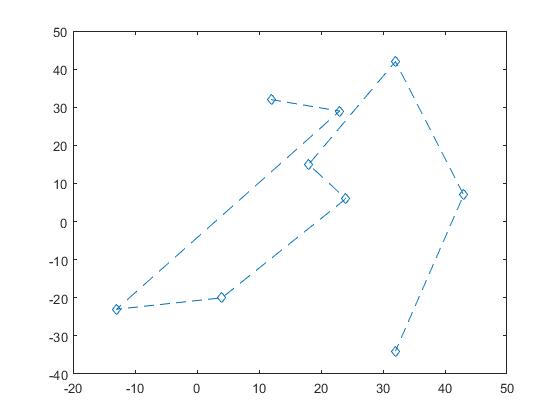}
\caption{Representation of Path in the form of way-Points}
\label{fig:waypoints}
\end{figure}

Given a pre-determined path, usually stored in the form of data points, known as way-points, as shown in Figure~\ref{fig:waypoints}, the initial position of the unmanned system $P(x,y)$ and heading angle $\psi$, the objective is to determine the controlled angle that most closely tracks the given path. The path can be approximated in the form of straight lines between way-points $ W_i$ and $ W_{i+1} $~\cite{aguiar2008performance}.
Angle $\theta$ formed by this straight line between $W_i$ and $W_{i+1}$ is called a line of sight (LOS). The problem statement targets to follow one straight line segment and apply a path planning algorithm to follow the current segment. This algorithm can easily be generalized to trace the path determined by all the way-points.
\begin{figure}
 \centering
\includegraphics[width=0.8\linewidth,clip,keepaspectratio]{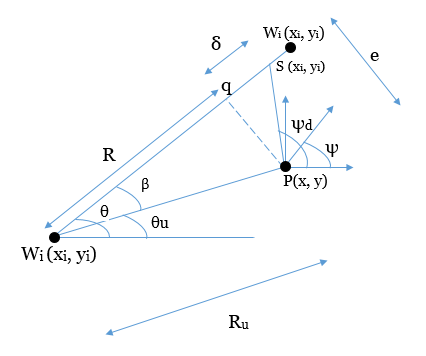}
\caption{Initial Geometry of the straight-line path defined by waypoints $W_i$ and
$W_{i+1}$ that needs to be followed by the unmanned system located at point P}
\label{fig:wp_geometry}
\end{figure}
Figure~\ref{fig:wp_geometry} shows the initial geometry of the problem. Distance $e$ from the unmanned system to the path is called a cross-track error. Let $\theta$ be the desired angle of the path. In addition to minimizing the cross-track error $e$, the unmanned system must minimize the heading error $|\theta-\psi|$ where $|~.~|$ represents the absolute value.
The objective of the path following algorithm reduces to have $d\xrightarrow[]{} 0$ and $|\psi \xrightarrow[]{} 0|$ as the mission time $t\xrightarrow[]{} \infty$.
Formally control problem can be defined as follows.
\subsection{Control Problem}
Let unmanned system is supposed to trace a pre-determined path $F(\alpha) \in \mathbb{R}^2$ which is parametrized by $\alpha \in \mathbb{R}^2$. The desired speed is given by $v_p(\alpha)\in \mathbb{R}$ and let $F(\alpha)$ be sufficiently smooth bounded derivative. The objective is to design a feedback control law such that the closed loop signals are bounded, $||p(t) - F(\alpha(t))||$, ~$p(t)$ being position function of unmanned system, converges to neighbourhood of the origin and the rate of change of error $|\alpha'(t) - v_p(\alpha(t))| < \varepsilon $ for $\varepsilon > 0$.

\section{Path Following Algorithm}
Assume that unmanned system maintains constant speed and is restricted in the x-y plane and their is no disturbance from wind, waves and any other external sources. Under this conditions kinematics of the unmanned system is given by~\cite{dubins1957curves, rhee2010tight}:
\spliteq{
v_x = v_acos(\psi)\\
v_y = v_asin(\psi)\\
\psi' = K(\psi_d -\psi)\\
u = max(\psi'*v_a)}

For the purpose of simulation in MATLAB\textsuperscript{\textregistered}, speed is assumed to be $v_a = 25m/s$. The maximum limit condition on $u$ in the simulation is to limit the maximum lateral acceleration. $K$ is the controller gain which is adjusted to find the stability point. The algorithm used to control the path is called \textbf{carrot-chasing algorithm}, imagining a rabbit chasing a moving carrot,  which employs the concept of virtual target point VTP. The Unmanned system updates its heading towards VTP. As time moves forward, the unmanned system moves toward the path, and asymptotically follows the path. In this algorithm, to follow the straight line segment defined by the waypoint $W_i$ and $W_{i+1}$, there are three main steps:
\begin{enumerate}
\item Determine the cross-track error $e$.
\item Update the location of Virtual Target Point $S$
\item Update $\psi_d$ and $u$
\end{enumerate}

\begin{figure}[htpb]
 \centering
\includegraphics[trim={2.0cm 2.0cm 2.0cm 1.4cm},clip,width=0.8\linewidth]{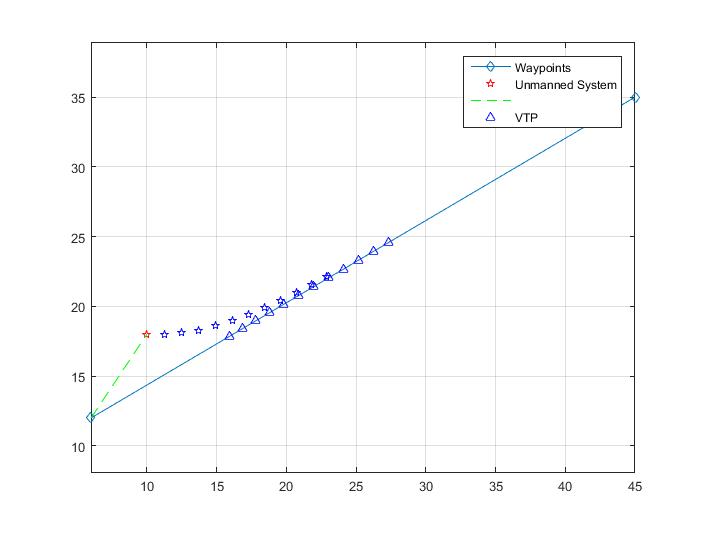}
\caption{$K = 1$, $\delta = 5 $}
\label{fig:sim1}
\end{figure}

\begin{figure}[htpb]
 \centering
\includegraphics[trim={2.0cm 2.0cm 2.0cm 1.4cm},width=0.8\linewidth]{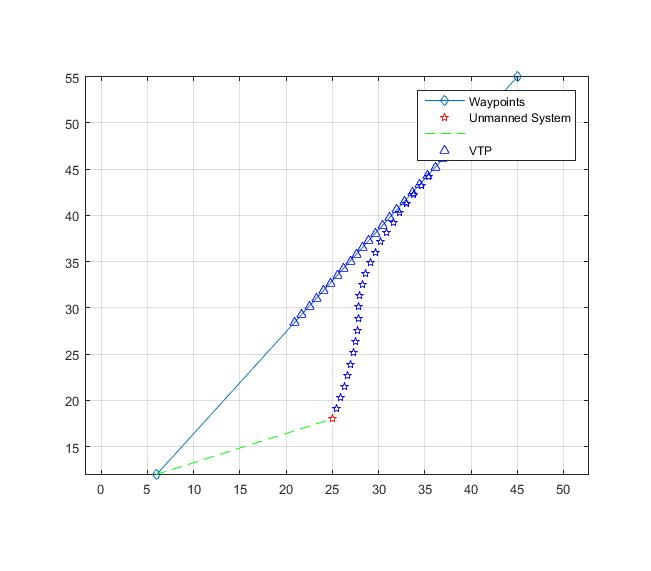}
\caption{$K = 3.5$, $\delta = 5 $}
\label{fig:sim2}
\end{figure}

In Figure~\ref{fig:wp_geometry}, the unmanned system is at location $P(x,y)$ with heading $\psi$. Projection of $P(x,y)$ on the line of sight is $q$ at a distance $R$ from $W_i$. The projection distance is cross-track error $e$. $S(x_i, y_i)$ is the VTP at distance $\delta$ from $q$. The control input in the algorithm uses a proportional controller with gain $K > 0$. For the purpose of simulation parameter $K = 0.5$ and $\delta = 5.0 $. With low value of $\delta$ the unmanned system quickly moves toward the path, but as $\delta$ is increased, cross-track error reduces very slowly because the unmanned system is slow in its asymptotic approach to the path. As the gain $K$ is increased, it directly moves to towards the path but for very high ( $K > 30$), poles of the transfer function shift to the right-hand plane, making the system unstable. We can write the overall carrot-chasing algorithm  as given in Section~\ref{sec:algorithm} and a few simulation results are provided in Figures~\ref{fig:sim1} and~\ref{fig:sim2}.
\subsection{Algorithm }
\label{sec:algorithm}
\begin{figure}[htpb]
 \centering
\includegraphics[trim={2.0cm 2.0cm 2.0cm 1.4cm},width=0.8\linewidth]{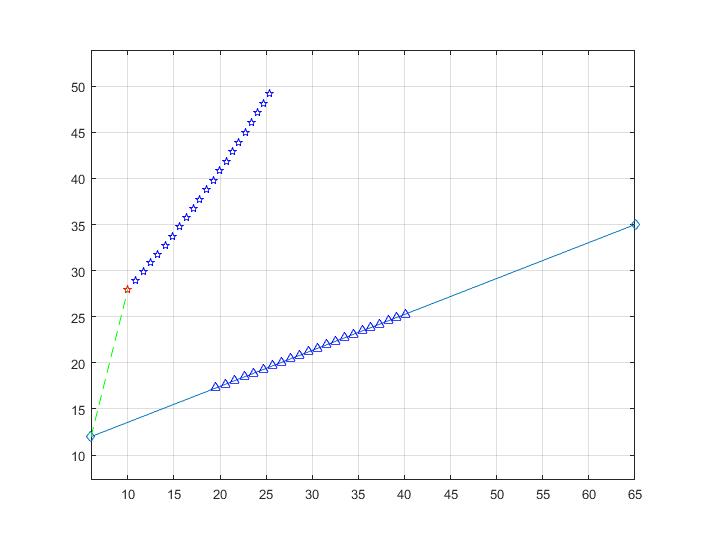}
\caption{K = 3.5, $\delta = 5 $}
\label{fig:sim3}
\end{figure}

\begin{figure}
 \centering
{\includegraphics[width=3.5in,clip,keepaspectratio]{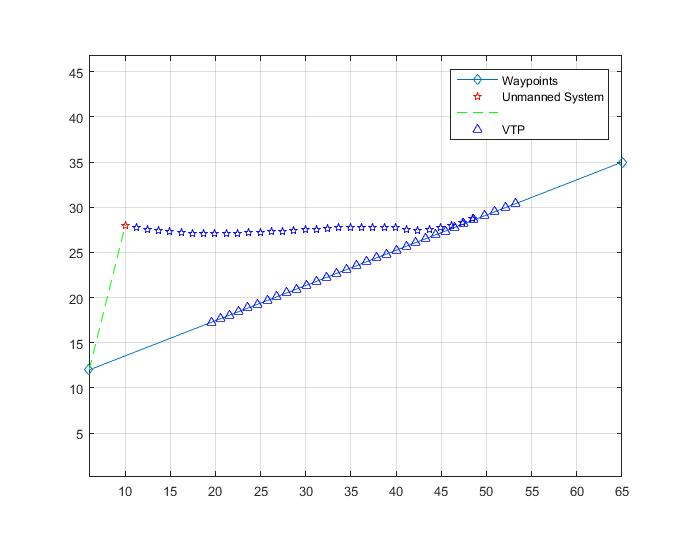}}
\caption{$K_1 = 0.5$, $K_2 = 35$, $\delta = 5 $}
\label{fig:sim4}
\end{figure}

\begin{enumerate}
\item Initialize $W_i(x_i, y_i), W_{i+1}(x_{i+1}, y_{i+1})$
\item Initialize $P = (x, y), \psi, \delta, v_a, k$
\item $ R_u = || W_i -P ||, \theta = tan^{-1}((y_{i+1} - y_i)/(x_{i+1} - x_i))$ 
\item $\theta_u = tan^{-1}((y - y_i)/(x - x_i)), \beta = \theta - \theta_u$
\item $R = Rcos(\beta)$
\item $S(x_t, y_t) = ((R+\delta)cos(\theta),(R+\delta)sin(\theta))$
\item $\psi_d = tan^{-1}(y_t - y/x_t - x)$
\item $u = max(K*(\psi_d - \psi)*v_a)$
\end{enumerate}
When the line of sight is less than 45 degrees, the control law given by Algorithm in Section~\ref{sec:algorithm} doesn't converge towards the path, as shown in Figure~\ref{fig:sim3}. In such case, a different control law is proposed which is as follows:
\spliteq{u = \max(K_1(\psi_d - \psi)v_a + K_2e)}
where $e$ is cross-track error. With $K_1 = 0.5$ and $K_2 = 35$, configuration shown in Figure~\ref{fig:sim3} converges to one shown in Figure~\ref{fig:sim4}.

\section{Conclusion}

The work presented in this project employs a very simple yet very elegant proportional controller to control the heading angle of the unmanned system to trace the pre-determined path. But definitely, it doesn't take into consideration the dynamics of the unmanned system and the influence of external environments such as gravity, air pressure, signal noise, and velocity component in the z-direction. It also assumes that speed is constant which is not always the case with those kinds of systems. Despite all these limitations, the rabbit-chasing algorithm can be extended to 3-dimensional coordinates. With more sophisticated controllers such as Linear Quadratic Regulator~\cite{dorf1998modern} and Nonlinear Guidance Law, the objective can be attained with a more relaxed system definition. Future work in this direction should consider the dynamics of the system and variability of parameters

\bibliographystyle{plain}
\bibliography{IEEEabrv,main.bib}

\section*{MATLAB Code}

%\begin{mdframed}[linecolor=black, topline=true, bottomline=true,
%  leftline=false, rightline=false, backgroundcolor=white!90!red]
%    \inputminted[fontsize=\footnotesize, linenos, frame=lines]{matlab}{CarrotChasing.m}
%\end{mdframed}
\footnotesize{
\begin{lstlisting}[style=matlab]

%Initialization
Va = 25; %m/s
phi = 0.9;
figure;
Vx = Va*cos(phi);
Vy = Va*sin(phi);
Wa = [6 65]; %x coordinates of waypoints
Wb = [12 35]; %y coordinates of waypoints
Px = 10; Py = 28;
P = [Px Py];
hold off;
plot(Wa, Wb,'-d');
%axis([0 60 0 60]);
axis equal;
grid on;

Ru = sqrt((Wb(1)-Py)^2 + (Wa(1) -Px)^2);
theta = abs(atan((Wb(2) - Wb(1))/...
    (Wa(2) - Wa(1))));
thetaU = abs(atan((Py - Wb(1))/...
    (Px - Wa(1))));
beta = abs(theta - thetaU);

R = Ru*cos(beta);
e = Ru*sin(beta);
hold on;
plot(Px,Py,'pr');
plot([Wa(1) Px],[Wb(1) Py],'--g');

delta = 5;
xt = Wa(1) + (R+delta)*cos(theta);
yt = Wb(1) + (R+delta)*sin(theta);
plot(xt,yt,'^b');

K =0.5;
K2 = 35;
while abs(e) > 0.2
    t = 0.05;
    
    phiD = abs(atan((yt -Py)/(xt-Px)));
    u = K*(phiD - phi)*Va - K2*e;
    %u = K*(phiD - phi)*Va;
    if u > 1
        u = 1;
    end
    phi = phiD;
    
    Vy = Va*sin(phi) + u*t;
    Vx = sqrt(Va*Va - Vy*Vy);
    Px = Px + Vx*t;
    Py = Py + Vy*t;
    t = t + 0.1;
    drawnow;
    plot(Px,Py,'pb');
    Ru = sqrt((Wb(1)-Py)^2 + (Wa(1) -Px)^2);
    thetaU = abs(atan((Py - Wb(1))/...
        (Px - Wa(1))));
    beta = abs(theta - thetaU);

    R = Ru*cos(beta);
    e = Ru*sin(beta);
    %plot([Wa(1) Px],[Wb(1) Py],'--g');
    xt = Wa(1) + (R+delta)*cos(theta);
    yt = Wb(1) + (R+delta)*sin(theta);
    plot(xt,yt,'^b');
    pause(1);
end

legend('Waypoints','Unmanned System','','VTP');
\end{lstlisting}
}

\end{document}